\documentclass[10pt,twocolumn,letterpaper]{article}

\usepackage{cvpr}              

\usepackage[dvipsnames]{xcolor}

\DeclareMathOperator*{\argmin}{arg\,min}

\definecolor{cvprblue}{rgb}{0.21,0.49,0.74}
\usepackage[pagebackref,breaklinks,colorlinks,citecolor=cvprblue]{hyperref}
\usepackage[accsupp]{axessibility}

\def\paperID{9135} 
\def\confName{CVPR}
\def\confYear{2024}

\title{Privacy-Preserving Face Recognition Using Trainable Feature Subtraction}

\author{
Yuxi Mi$^{1}$\thanks{Authors contributed equally to this paper.}\quad
Zhizhou Zhong$^{1}\footnotemark[1]$\quad
Yuge Huang$^{2}$\thanks{Corresponding authors.}\quad
Jiazhen Ji$^{2}$\quad
Jianqing Xu$^{2}$\\
Jun Wang$^{3}$\quad
Shaoming Wang$^{3}$\quad
Shouhong Ding$^{2}$\quad
Shuigeng Zhou$^{1}\footnotemark[2]$
\\
$^{1}$ Fudan University \quad
$^{2}$ Youtu Lab, Tencent \quad
$^{3}$ WeChat Pay Lab33, Tencent
\\
{\tt\small yxmi20@fudan.edu.cn, zzzhong22@m.fudan.edu.cn, sgzhou@fudan.edu.cn} \\
{\tt\small \{yugehuang, royji, joejqxu, ericshding\}@tencent.com} \\
{\tt\small \{earljwang, mangosmwang\}@tencent.com}
}

\begin{document}
\maketitle
\begin{abstract}

The widespread adoption of face recognition has led to increasing privacy concerns, as unauthorized access to face images can expose sensitive personal information. This paper explores face image protection against viewing and recovery attacks. Inspired by image compression, we propose creating a visually uninformative face image through feature subtraction between an original face and its model-produced regeneration. Recognizable identity features within the image are encouraged by co-training a recognition model on its high-dimensional feature representation. To enhance privacy, the high-dimensional representation is crafted through random channel shuffling, resulting in randomized recognizable images devoid of attacker-leverageable texture details. We distill our methodologies into a novel privacy-preserving face recognition method, MinusFace. Experiments demonstrate its high recognition accuracy and effective privacy protection. Its code is available at \url{https://github.com/Tencent/TFace}.

\vspace{-4mm}
\end{abstract}    
\section{Introduction}
\label{sec:intro}

Face recognition (FR) is a biometric way to identify persons through their face images. It has seen prevalent methodological and application advancements in recent years. Currently, considerable parts of FR are implemented as online services to overcome local resource limitations: Local clients, such as cell phones, outsource captured face images to an online service provider. Using its model, the provider extracts identity-representative templates from the face images and matches them with its database.

It has been common sense that face images are sensitive biometric data and should be protected. Increasing regulatory demands~\cite{voigt2017eu} call for privacy-preserving face recognition (PPFR), to avoid leakage of face images during the outsourcing. They attempt to ensure that the faces' appearances are both \textit{visually concealed} from inadvertent view by third parties and \textit{difficult to recover} by deliberate attackers. 

State-of-the-art (SOTA) PPFR primarily employs two approaches: Cryptographic methods protect face images with encryption or security protocols. Recently, transform-based methods have gained popularity due to their low latency and budget-saving computational costs. They convert images into protective representations by minimizing visual details, rendering them safe to share.

\begin{figure}[tbp]
  \centering
   \includegraphics[width=\linewidth]{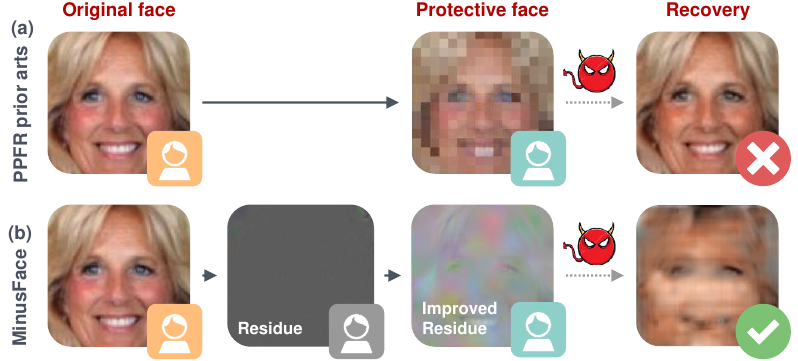}
   \caption{Comparison between SOTAs and MinusFace. (a) SOTAs gradually remove the most visually informative features. Inadequacy of removal can result in successful recovery, which undermines privacy. (b) MinusFace first obtains a \textit{fully} visually uninformative residue representation, then improves its recognizability. It exhibits better privacy protection than all SOTAs.} 
   \label{fig:paradigm}
   \vspace{-5mm}
\end{figure}

Transform-based methods yet face a persistent challenge in balancing accuracy and privacy. In face images, the recognizable \textit{identity features} and appearance-revealing \textit{visual features} are closely intertwined. To achieve privacy while preserving optimal recognizability, prior arts invest significant efforts to locate and minimize the most visually informative feature components while retaining the rest. They commonly employ either a heuristic or adversarial training approach: For instance, some~\cite{wang22ppfrfd,DBLP:conf/mm/MiHJLXDZ22,DBLP:conf/eccv/JiWHWXDZCJ22,mi2023privacy} turn face images into frequency domain and heuristically prune the most human-perceivable frequency channels. Others exploit deep steganography~\cite{yuan2022pro} or cyclically add adversarial noise~\cite{wang2023privacy}. While these methods succeed in concealing faces from human inspection, they can be \textit{largely susceptible to recovery attacks}~\cite{dosovitskiy2016inverting,he2019model,mai2018reconstruction}. Their challenge lies in ensuring an adequate removal of visual features, as subtle features may remain, providing attackers with potential leverage.

\begin{figure}[tbp]
  \centering
   \includegraphics[width=\linewidth]{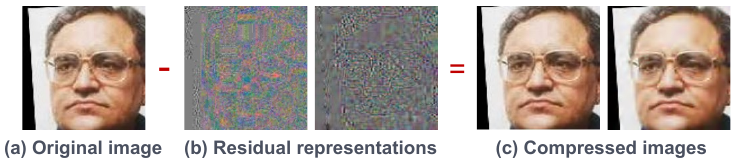}
   \caption{Examples of image compression. Subtle details like texture are removed from (a) the original image to obtain (c) the compressed ones. The removed (b) residual representations are visually uninformative, yet carry descriptive features of the origin.}
   \label{fig:compression}
   \vspace{-4mm}
\end{figure}

This paper advocates a novel approach to more effectively minimize visual features, drawing inspiration from image compression. Image compression reduces image size while preserving fidelity by discarding subtle features such as texture details and color variations. The paper observes that the discarded features, \ie, the \textit{residue} between original and compressed images, exhibit properties closely aligned with the desired protective face representation: They are both visually uninformative and preserve descriptive features of the original image, as shown in~\cref{fig:compression}.

Emulating the production of discarded features, this paper introduces a trainable \textit{feature subtraction} strategy to craft protective representations. In this approach, a generative model is first trained to faithfully produce a regeneration of the original face, where the regenerated face simulates a compressed image. The residue between the original and regenerated faces is expected to be devoid of visual features if the model is well-optimized. It is later exploited to produce a protective representation. To retain recognizability within the residue, a recognition model is co-trained, taking the residue as input to learn identity features.

Two techniques are subsequently proposed to enhance both the recognizability and privacy of the residue. To address specific training constraints of the FR model (detailed in~\cref{subsec:high-dim-constrain}), the residue is generated as \textit{high-dimensional representations} instead of spatial images, enabling better preservation of identity features. Privacy is heightened through \textit{random channel shuffling}, which obscures facial texture signals and increases randomness to hinder recovery attacks. The shuffled high-dimensional residue is ultimately mapped back as a spatial image, serving as the protective representation. The methodology is concretized into a novel PPFR framework, MinusFace. \Cref{fig:paradigm} compare it with SOTA prior arts in paradigm. Experimental results show that MinusFace achieves high recognition accuracy and better privacy protection than SOTAs.

This paper presents three-fold contributions:

\begin{itemize}[leftmargin=2em]
    \item It introduces \textit{feature subtraction}, a new methodology to generate protective face representation, by capturing residue between an original image and its regeneration.
    \item It proposes two specific techniques, high-dimensional mapping and random channel shuffling, to ensure recognizability and accuracy for the residue.
    \item It presents a novel PPFR method, MinusFace. Experimental results demonstrate its high recognition accuracy and superior privacy protection to SOTAs.
\end{itemize}

\section{Related work}
\label{sec:related-work}

\subsection{Face recognition}

Current FR systems identify persons by comparing their face templates, \ie, one-dimensional feature embeddings. The service provider trains a convolutional neural network (CNN) to extract templates from face images. With angular-margin-based losses~\cite{DBLP:conf/cvpr/DengGXZ19,DBLP:conf/cvpr/WangWZJGZL018,DBLP:conf/cvpr/LiuWYLRS17,huang2020curricularface,dan2023transface}, the templates are encouraged to have large inter-identity and small intra-identity discrepancies that facilitate recognition.

\subsection{Privacy-preserving face recognition}

Many approaches have been proposed to protect face privacy~\cite{DBLP:journals/ijon/WangD21a,DBLP:journals/tifs/MedenRTDKSRPS21,DBLP:journals/corr/abs-2206-10465}. We divide them into two categories.

\noindent \textbf{Cryptographic methods} perform recognition on encrypted face images. To allow necessary computations in the cipher space, many prior arts employ homomorphic encryption~\cite{DBLP:conf/icisc/SadeghiSW09,DBLP:conf/pet/ErkinFGKLT09,DBLP:conf/icml/Huang0LA20,ibarrondo2023grote,yang2023review,huang2023efficient} or secure multiparty computation~\cite{DBLP:journals/iotj/MaLLMR19,DBLP:conf/ithings/YangZLLL18,DBLP:journals/soco/XiangTCX16,ren2022privacy} to extract encrypted features and calculate their pair-wise distances. Others leverage different crypto-primitives including matrix encryption~\cite{DBLP:conf/acmturc/KouZZL21}, one-time-pad~\cite{DBLP:conf/apccas/Ergun14a}, functional encryption~\cite{DBLP:conf/pkc/AbdallaBCP15}, and locality-sensitive hashing~\cite{zhao2023priface,gao2023privacy}. These methods, however, mostly bear high latency and heavy computational overheads.

\noindent \textbf{Transform-based methods} convert face images into protective representations that cannot be directly viewed. Pioneering arts obfuscate the image by adding crafted noise~\cite{DBLP:journals/finr/ZhangHXGY20,DBLP:journals/compsec/ChamikaraBKLC20,DBLP:journals/ijon/LiWL19b,DBLP:conf/hotedge/X18,wen2022identitydp}, performing clustering~\cite{DBLP:conf/socpar/HondaOUN15}, or extracting coarse representations~\cite{DBLP:conf/autoid/KevenaarSVAZ05,DBLP:conf/mlsp/ChanyaswadCMK16,DBLP:conf/www/MireshghallahTJ21,sun2022privacy}. Some regenerate the faces’ features to obtain different visual appearances using autoencoders~\cite{DBLP:conf/icb/MirjaliliRNR18,tang2022gender}, adversarial generative networks~\cite{DBLP:conf/btas/MirjaliliRR18,DBLP:journals/ijon/LiWL19b,boutros2022sface}, and diffusion models~\cite{huang2023collaborative,boutros2023idiff}. However, these methods suffer from compromised recognition accuracy as the obfuscation and regeneration often indiscriminately degrade the faces’ visual and identity features. Recent methods locate and modify the images’ most visually informative components. \cite{wang22ppfrfd,DBLP:conf/mm/MiHJLXDZ22,DBLP:conf/eccv/JiWHWXDZCJ22,mi2023privacy} transform images to the frequency domain, where human-perceivable low-frequency channels are pruned. \cite{yuan2022pro} uses deep steganography to conceal the face under distinct carrier images and aligns identity features via contrastive loss. \cite{wang2023privacy} generates protective features by cyclically adding adversarial noise to sensitive signals. These methods visually conceal face appearance quite successfully and maintain decent recognition accuracy. However, we later experimentally show that they can be vulnerable to recovery attacks.

\section{Methodology}
\label{sec:method}

This section describes our proposed MinusFace. The name comes from the key methodology to produce the protective representation, by subtracting between the original face and its regeneration, \ie, the ``\textit{minus}''. 
We begin by learning a visually uninformative representation in~\cref{subsec:toy-framework} via feature subtraction. In~\cref{subsec:high-dim-constrain}, we improve the representation in high-dimension to let it preserve identity features. In~\cref{subsec:rand-shuffle}, we further address its privacy and to produce the final protective representation.

\subsection{Motivation}
\label{subsec:motivation}

The general goal of transform-based PPFR is to design a privacy-preserving transformation $F$ that converts any original face image $X$ to a \textit{protective representation} $X_p=F(X)$. In prior arts, $X_p$ can be concretized as a spatial image~\cite{DBLP:journals/compsec/ChamikaraBKLC20,DBLP:conf/icml/Huang0LA20,DBLP:journals/tifs/TsengW20,wang2023privacy} or high-dimensional feature channels~\cite{wang22ppfrfd,DBLP:conf/mm/MiHJLXDZ22,DBLP:conf/eccv/JiWHWXDZCJ22,mi2023privacy,wang2023privacy}. Either way, we expect $X_p$ to \textit{preserve identity features} and \textit{minimize visual features}. 

Our approach is enlightened by image compression, a technique that reduces image size while preserving fidelity. Specifically, lossy image compression methods~\cite{theis2017lossy,DBLP:journals/cacm/Wallace91,skodras2001jpeg} exploit human perceptual insensitivity to discard subtle features like texture details or color variations. Interestingly, the discarded features possess the desired properties for our $X_p$: Since they are considered insignificant to image fidelity, they should be visually uninformative; otherwise, the compression would be too lossy. Meanwhile, viewing these features as the \textit{residual representation}, denoted as $R$, between the original and compressed image, they still contain the image's clues. If the residue $R$ can be utilized for recognition, the compression factually provides us with a natural $X_p$ that is both visually indiscernible and recognizable.

\Cref{fig:compression}(b) demonstrates the residual representations of an example face image under JPEG~\cite{DBLP:journals/cacm/Wallace91} and JPEG 2000~\cite{skodras2001jpeg} compression standards. We can observe feature clues from both residual representations. Regretfully, we find directly using them for face recognition ends up quite ineffective because their features are not specifically manufactured to keep the identity. In fact, they behave more like random noise from the perception of FR models.

\subsection{Feature subtraction: minimize visual features}
\label{subsec:toy-framework}

\begin{figure}[tbp]
  \centering
   \includegraphics[width=\linewidth]{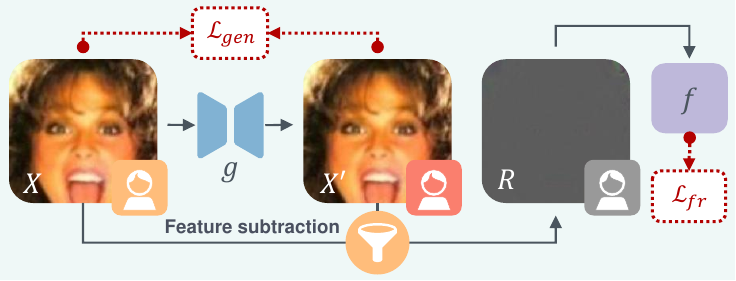}
   \caption{The core idea of MinusFace. Imitating image compression, a visually uninformative residue $R$ is generated from \textit{feature subtraction}: the original face \textit{minus} its regeneration. $R$ is also optimized with an FR model to preserve identity features.} 
   \label{fig:toy-pipeline}
   \vspace{-4mm}
\end{figure}

Imitating image compression, we can produce a residual representation $R$ that is recognizable through a trainable \textit{feature subtraction} strategy: To minimize visual features, we train a model that regenerates a face image $X'$ taking the original face $X$ as input. It simulates the compression process. We produce $R$ as the subtraction between $X$ and $X'$, \ie their \textit{minus}, which should be visually uninformative if the regeneration is successful. Unlike image compression, crucially, we meanwhile train an FR model that tries to recognize $R$. By balancing the training of two models, $R$ should also preserve identity features once the FR model is optimized. Such an $R$ hence may serve as our protective representation $X_p$. \Cref{fig:toy-pipeline} demonstrates our idea. 

We first concretize the minimization of visual features. Specifically, let $g$ be a generative model (\textit{wlog.}, a CNN autoencoder). We regenerate a face image from the original face as $X'=g\left(X\right)$. To make $X'$ visually close to $X$, we employ $l_1$-norm as the model's objective:

\begin{equation}
    \label{eq:vanilla-l1}
    \mathcal{L}_{gen}= \left\|X,X'\right\|_{1}.
\end{equation}

\noindent Prior studies~\cite{DBLP:conf/mm/MiHJLXDZ22,mi2023privacy} suggest optimizing~\cref{eq:vanilla-l1} is trivial provided the original face is not further obfuscated, which is our case. Therefore, we can confidently obtain a regeneration with high fidelity, $X' \approx X$. We produce the residue as their subtraction, $R=X-X'$.

As earlier discussed, prior arts invest huge efforts in removing the most visually informative features from $X_p$. However, their removals of features are often inadequate, resulting in unsatisfactory privacy. Leveraging feature subtraction, we efficiently transform the feature-minimizing objective of $X_p$ to the feature-maximizing goal of $X'$, which is easier to quantify: Instead of explicitly removing $R$'s visual features, since~\cref{eq:vanilla-l1} can be rewritten as

\begin{equation}
    \label{eq:vanilla-l1-rewrite}
    \mathcal{L}_{gen}= \left\|X,X'\right\|_{1} = \left\|X-X',0\right\|_{1}=\left\|R\right\|_{1},
\end{equation}

\noindent we can expect $R$ to be visually uninformative simply by producing high-quality $X'$.

\subsection{Preserve identity features in high-dimension}
\label{subsec:high-dim-constrain}

\begin{figure*}[tbp]
  \centering
   \includegraphics[width=\linewidth]{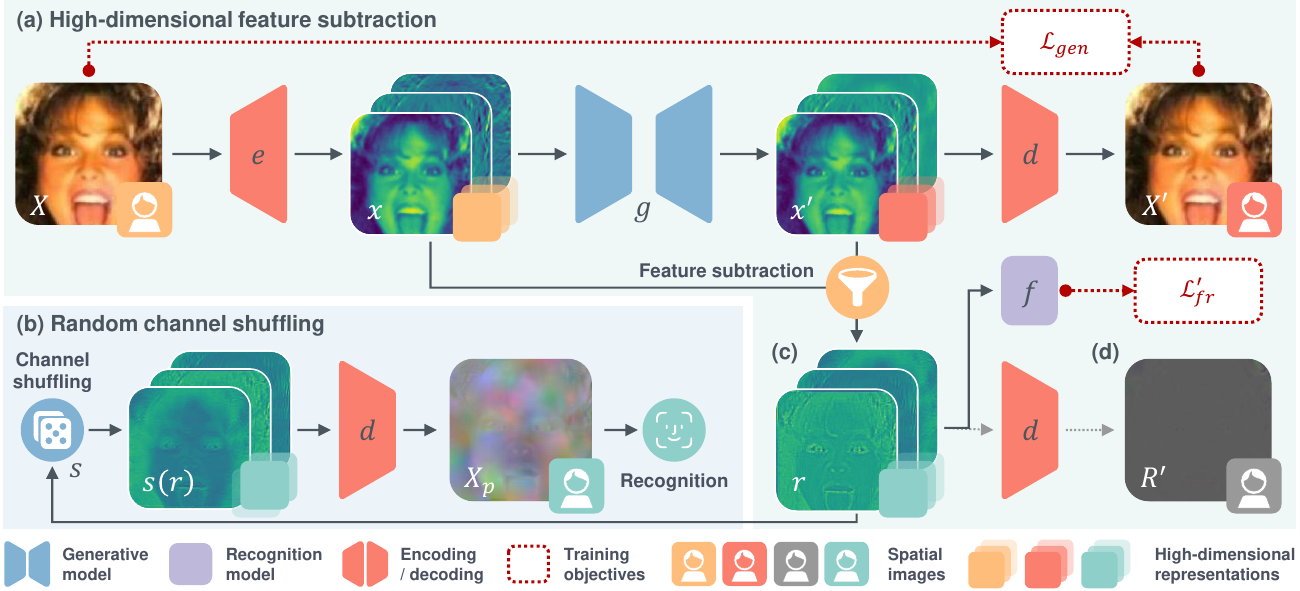}
   \caption{The MinusFace pipeline. (a) It centers around the idea of feature subtraction, where the protective representation$X_p$ is derived from the residue between the original face $X$ and its regeneration $X'$. Both regeneration and feature subtraction occur in high-dimension to preserve identity features within the trained residue $r$. (b) The residue $r$ further undergoes random channel shuffling and decoding to produce the protective representation $X_p$. (c-d) All face figures are experimentally obtained and illustrate their representations faithfully. } 
   \label{fig:pipeline}   
   \vspace{-4mm}
\end{figure*}

Next, we aim to obtain the identity features for $R$ to make it recognizable. As illustrated in~\cref{fig:toy-pipeline}, the most intuitive strategy is to incorporate an FR model $f$ that takes $R$ as input. Let $f$ be end-to-end trained with the generative model $g$, aiming to predict the face's identity $y$. Thus, $R$ should acquire identity features as long as $f$ is also optimized.

However, we find training $f$ can be challenging as it often ends up in poor convergence. We owe it to a slight drawback of feature subtraction: By optimizing~\cref{eq:vanilla-l1}, it in fact indiscriminately removes both visual and identity features, encouraging $R$ to be blank. In other words, feature subtraction is trading off recognizability for privacy.

We propose a strategy that circumvents the trade-off, inspired by the property of high-dimensional spaces. Specifically, high-dimensional spaces often contain \textit{significant redundancy of features}. If we map a spatial image $X\in \mathbf{M}$ to a high-dimensional representation $x\in \mathbf{N}$, we can expect $X$'s visual appearance to be described by very few of $x$'s components, \ie, the \textit{principal components}. The remaining features of $x$ can then be reorganized without changing $X$. Let $x,x'$ be the high-dimensional representations of $X, X'$, respectively. While feature subtraction enforces $X'\rightarrow X$, likely making the principal components of $x,x'$ identical, we can make a difference in their less visually descriptive and abundant remaining features. This allows us to produce non-blank high-dimensional residue $r=x-x'\neq 0$, which can carry identity features.

We establish a pair of differentiable, deterministic encoding $e:\mathbf{M}\rightarrow \mathbf{N}$ and decoding $d:\mathbf{N}\rightarrow \mathbf{M}$ mappings to handle the conversion between $\mathbf{M},\mathbf{N}$. As properties necessary for later discussions, we require $d,e$ together to be \textit{invertible} and $d$ alone to be \textit{homomorphic}, \ie,

\begin{equation}
    \label{eq:mapping-property}
    \begin{cases}
        d\left(e(a)\right) = a& \forall a,\\
        d(a_1 + a_2)= d(a_1)+d(a_2) & \forall a_1, a_2.
    \end{cases}
\end{equation}

\noindent Also inspired by image compression, we choose \textit{discrete cosine transform} (DCT) and its inverse (IDCT) as $d,e$, respectively. DCT is a linear transformation employed in JPEG~\cite{DBLP:journals/cacm/Wallace91} compression, that converts a $(3, H, W)$ image $X$ into a $(192, H, W)$ high-dimensional $x$. We provide further details in the supplementary material. We opt for DCT, \textit{wlog.}, for three main reasons: (1) It satisfies~\cref{eq:mapping-property}; (2) It produces $x$ that preserves $X$'s spatial structure and feature information: Study~\cite{DBLP:conf/cvpr/0007QSWCR20} shows models trained on $x$ achieve similar performance as those on $X$; (3) It produces an $x$ with 192 channels. The abundant channels later enhance privacy by shuffling their orders. Nonetheless, other $d,e$ may be chosen provided at least~\cref{eq:mapping-property} is satisfied.

Here, we describe producing $r$ via high-dimensional feature subtraction. \Cref{fig:pipeline}(a) shows its pipeline. Note that all face figures here are experimentally obtained from MinusFace and illustrate their representations faithfully.

Specifically, we begin by encoding the high-dimensional representation of face image $X$ as $x=e(X)$. Then, we regenerate $x'=g(x)$ using the model $g$, which is modified to accept a 192-channel input, and subsequently decode it into a spatial image as $X'=d(x')$. Similar to~\cref{eq:vanilla-l1}, $g$ is trained by minimizing the $l1$-norm between $X$ and $X'$.

Meanwhile, we can avoid a blank residue $r$ by performing feature subtraction in high-dimension: We obtain the residue as $r=x-x'\neq 0$ and train the FR model $f$ on $r$ to help it acquire identity features, as previously discussed. The FR model $f$ can be optimized using any SOTA FR loss; \textit{Wlog.}, we opt for the popular ArcFace loss~\cite{DBLP:conf/cvpr/DengGXZ19}:

\begin{equation}
    \label{eq:minus-arcface}
        \mathcal{L}_{fr}=l_{arc}(f(r),y).
\end{equation}

\noindent The overall training objective of MinusFace is the combination of~\cref{eq:vanilla-l1,eq:minus-arcface}, weighted by $\alpha,\beta$:

\begin{equation}
    \label{eq:minus-objective}
    \mathcal{L}_{minus}=\alpha\cdot\mathcal{L}_{gen}+\beta\cdot\mathcal{L}_{fr}.
\end{equation}

\noindent We experimentally find both loss terms are optimized smoothly, and the produced residue $r$ can be recognized by $f$ with high accuracy, later shown in~\cref{subsec:exp-abal}. Hence, by mapping $X$ into high-dimension, we can acquire an $r$ with identity features under feature subtraction. This satisfies our recognizability goal.

Before closing this section, further let $R'=d(r)$ be the decoding of $r$. Interestingly and crucially to the following discussion, we find $R'=R$, \ie, $R'$ equal to the spatial residue between $X$ and $X'$ that is \textit{always blank}. The blankness of $R'$ is contributed by the properties of $d,e$. Note that $X$ can be rewritten by~\cref{eq:mapping-property} as

\begin{equation}
    \label{eq:x-invertible}
    X = d(e(X)) = d(x).
\end{equation}

\noindent Combining~\cref{eq:vanilla-l1} with $d$'s homomorphism, it always holds 
\begin{equation}
    \label{eq:minus-l1-rewrite}
    \begin{aligned}
        \mathcal{L}_{gen} &= \left\|X-X'\right\|_{1} \\
            & = \left\|d(x)-d(x')\right\|_{1} =\left\|d(x-x')\right\|_{1} \\
            & = \left\|d(r)\right\|_{1} = \left\|R'\right\|_{1}.
    \end{aligned}
\end{equation}

\noindent In~\cref{fig:pipeline}(d), we exhibit sample $R'$ experimentally generated, which is indeed blank. We use $r$ and its mapping to a blank $R'$ as key tools to produce the final protective $X_p$.

\subsection{Random channel shuffling}
\label{subsec:rand-shuffle}

The previous section creates a recognizable residue $r$. It is important to highlight that $r$ cannot \textit{directly} function as $X_p$ since it lacks a guarantee of privacy: Feature subtraction only ensures removing visual features from $R'$, but not necessarily from $r$. As exhibited in~\cref{fig:pipeline}(c), subtle visual features in sample $r$ persist, compromising its privacy. 

To bridge the privacy gap, this section shows that a protective $X_p$ can be simply derived as \textit{perturbing then decoding} $r$, without further training. Specifically, we choose to perturb $r$ by randomly shuffling its channels. Let $r_{\Delta}=s(r;\theta)$ represents the shuffling of $r$, where the channel order is determined by a \textit{sample-wise} random seed $\theta$. Thus, $X_p=d(s(r;\theta))$ serves as our final protective representation. The process is illustrated in~\cref{fig:pipeline}(b). Following, we explain the motivations behind our design.

We first show shuffling will provably gain $X_p$ with recognizability. Recall that $r$ is primarily mapped to a blank $R'=d(r)\rightarrow 0$ devoid of any features, ensured by~\cref{eq:minus-l1-rewrite}. Introducing a slight perturbation $\Delta r$ as $r_{\Delta}=r+\Delta r$ plausibly results in a disrupted $R'_{\Delta}\neq R'$. Note that $R'_{\Delta}$ cannot be less informative than $R'$ as the latter is already blank of identity features. Conversely, it \textit{acquires} features from the perturbation $\Delta r$, according to $d$'s homomorphism:

\begin{equation}
    \label{eq:perturb}
    d(r+\Delta r)=d(r) +d(\Delta r)\rightarrow d(\Delta r),
\end{equation}

\noindent with believably $\left\|d(\Delta r)\right\|_{1}>0$ unless rare circumstances. Further to note that shuffling $r$' channels equals choosing

\begin{equation}
    \label{eq:shuffle}
    \Delta r = r - s(r;\theta) \neq 0.
\end{equation}

\noindent  Given that $r$ preserves the identity features of the face image $X$, we anticipate that its shuffle $s(r;\theta)$ and their subtraction $\Delta r$ will also be identity-descriptive of $X$. Consequently, $X_p$ is able to assimilate the identity features of $r$ through shuffling. In~\cref{subsec:exp-accuracy}, we experimentally validate that the learning of identity features is robust, as FR on $X_p$ attains satisfactory recognition accuracy.

\begin{figure}[tbp]
  \centering
   \includegraphics[width=\linewidth]{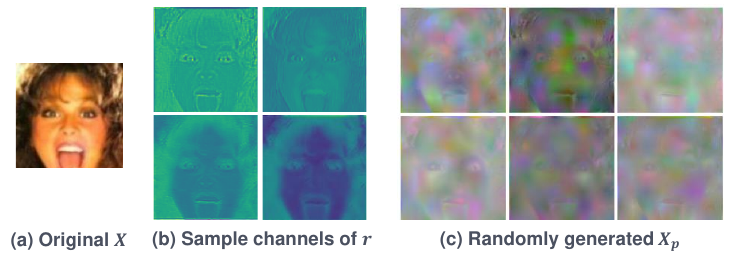}
   \caption{By randomly shuffling (b) channels of $r$, $192!$ distinct (c) $X_p$ can be generated from (a) the same $X$. We exhibit some channels and $X_p$. Different $X_p$ possess random texture patterns that obfuscate the recovery, by the nature of channel shuffling.}
   \label{fig:samples}
   \vspace{-3mm}
\end{figure}

\begin{table*}[tbp]
\centering
\begin{tabular}{lccccccccc}
\toprule
\textbf{Method}           & \textbf{Venue} & \textbf{PPFR} & \textbf{LFW} & \textbf{CFP-FP} & \textbf{AgeDB} & \textbf{CPLFW} & \textbf{CALFW} & \textbf{IJB-B} & \textbf{IJB-C} \\
\midrule
ArcFace~\cite{DBLP:conf/cvpr/DengGXZ19}          & CVPR '19      & No            & 99.77        & 98.30           & 97.88          & 92.77          & 96.05          & 94.13          & 95.60          \\
\midrule
PEEP~\cite{DBLP:journals/compsec/ChamikaraBKLC20}             & CS '20        & Yes           & 98.41        & 74.47           & 87.47          & 79.58          & 90.06          & 5.82           & 6.02           \\
InstaHide~\cite{DBLP:conf/icml/Huang0LA20}        & ICML '20      & Yes           & 96.53        & 83.20           & 79.58          & 81.03          & 86.24          & 61.88          & 69.02          \\
Cloak~\cite{DBLP:conf/www/MireshghallahTJ21}            & WWW '21       & Yes           & 98.91        & 87.97           & 92.60          & 83.43          & 92.18          & 33.58          & 33.82          \\
\midrule
PPFR-FD~\cite{wang22ppfrfd}          & AAAI '22      & Yes           & 99.69        & 94.85           & 97.23          & 90.19          & 95.60          & 92.93          & 94.07          \\
DCTDP~\cite{DBLP:conf/eccv/JiWHWXDZCJ22}            & ECCV '22      & Yes           & 99.77        & 96.97           & 97.72          & 91.37          & 96.05          & 93.29          & 94.43          \\
DuetFace~\cite{DBLP:conf/mm/MiHJLXDZ22}         & MM '22        & Yes           & 99.82        & 97.79           & 97.93          & 92.35          & 96.10          & 93.66          & 95.30          \\
PartialFace~\cite{mi2023privacy}      & ICCV '23      & Yes           & 99.80        & 97.63           & 97.79          & 92.03          & 96.07          & 93.64          & 94.93          \\
ProFace~\cite{yuan2022pro}          & MM '22        & Yes           & 98.27        & 93.77           & 92.81          & 88.17          & 93.20          & 69.39          & 72.96          \\
AdvFace~\cite{wang2023privacy}          & CVPR '23      & Yes           & 98.45        & 92.21           & 92.57          & 83.73          & 93.62          & 70.21          & 74.39          \\
\textbf{MinusFace} & (ours)              & Yes           & 99.78            & 96.92               & 97.57              & 91.90              & 95.90              & 93.37              & 94.70       \\
\bottomrule
\end{tabular}
\caption{The performance comparison among MinusFace, an unprotected baseline, and PPFR SOTAs on face verification and identification tasks. MinusFace achieves on-par ($\pm1\%$) performance with the best frequency-based SOTAs and outperforms the others.}
\label{tab:comp-sota}
\vspace{-4mm}
\end{table*}

We opt for random channel shuffling over other perturbations as it helps minimize privacy costs. Through perturbation, $X_p$ is bound to unintentionally recover some visual features from $r$ due to the intertwining of visual and identity features. In this context, shuffling demonstrates two-fold privacy benefits: \textit{natural obfuscation} of visual features and \textit{introduction of randomness} to $X_p$'s representations.

To explain the natural obfuscation, we closely examine the sample channels of $r$ in~\cref{fig:samples}(b). We find these channels reveal consistent signals in structure (\eg, positions for eyes and noses) but diverse ones in texture (\eg, color depths). This phenomenon arises from the use of CNN-based $g$ (and spatial-preserving $d,e$), wherein CNNs inherently preserve the spatial relations of images and generate distinct channel-wise signals through various convolutional kernels. Existing studies~\cite{DBLP:conf/mm/MiHJLXDZ22,mi2023privacy,huang2023improving,si2023freeu,chen2022geometry} suggest that structural signals play a pivotal role in FR models, while generative models (say, \textit{the attacker's recovery model}) rely on both structural and texture signals. In our scenario, the structural signals consistent across channels prove more resistant to shuffling than the texture features with channel-wise variations. \Cref{fig:samples}(c) illustrates different samples of $X_p$ generated from the same $X$ under varied $\theta$. These samples exhibit very subtle facial contours similar to that of $X$, facilitating recognition. In contrast, their facial texture is transformed into meaningless color patches. This outcome of shuffling factually allows us to \textit{selectively} obfuscate most visual features while preserving identity features, achieving an improved privacy-accuracy trade-off.

Privacy is further enhanced through the randomness of produced $X_p$. A successful recovery attack~\cite{dosovitskiy2016inverting,he2019model,mai2018reconstruction} necessitates training the attack model on consistent representations. Recall that $r$ is a high-dimensional representation with a shape of $(192, H, W)$. Randomly shuffling its channels can produce $192!$ different $X_p$ with random textures for the same $X$. The attacker can neither learn from $X_p$ with random textures nor determine the seed $\theta$ for a specific $X_p$. Results in~\cref{subsec:exp-recovery,subsec:exp-recovery-2} show MinusFace completely nullifies SOTA recovery attacks.

\subsection{Summary}
\label{subsec:summary}

To deploy MinusFace, the service provider first trains $f,g$ under~\cref{eq:minus-objective} that produces $r$. It discards $f$, as $f$ does not serve as the final FR model. It shares frozen $g$ with its clients. Capturing $X$, the clients obtain protective representation $X_p=F(X)$ with random $\theta$, outsourcing it to the provider. The provider recognizes $X_p$ on a newly trained FR model $f_p$. The same FR result is expected regardless of specific $\theta$. The final privacy-preserving transformation of MinusFace is $F=d(s(r;\theta))$, where $r=e(X)-e(g(X))$.
\section{Experiments}
\label{sec:exp}

\subsection{Experimental setup}
\label{subsec:exp-setup}

\noindent \textbf{Model and dataset.} We employ a U-Net~\cite{DBLP:conf/miccai/RonnebergerFB15} autoencoder with reduced scale as $g$, and IR-50~\cite{DBLP:conf/cvpr/HeZRS16} models as $f,f_p$. Training is carried out on the MS1Mv2~\cite{DBLP:conf/eccv/GuoZHHG16} dataset, which possesses 5.8M face images. We carry out evaluations on 5 regular-size datasets, LFW~\cite{LFWTech}, CFP-FP~\cite{DBLP:conf/wacv/SenguptaCCPCJ16}, AgeDB~\cite{DBLP:conf/cvpr/MoschoglouPSDKZ17}, CPLFW~\cite{CPLFWTech} and CALFW~\cite{DBLP:journals/corr/abs-1708-08197}. We also use 2 large-scale datasets, IJB-B~\cite{DBLP:conf/cvpr/WhitelamTBMAMKJ17} and IJB-C~\cite{DBLP:conf/icb/MazeADKMO0NACG18}. We leave further experimental and training setup to the supplementary material.

\begin{figure*}[tbp]
  \centering
   \includegraphics[width=\linewidth]{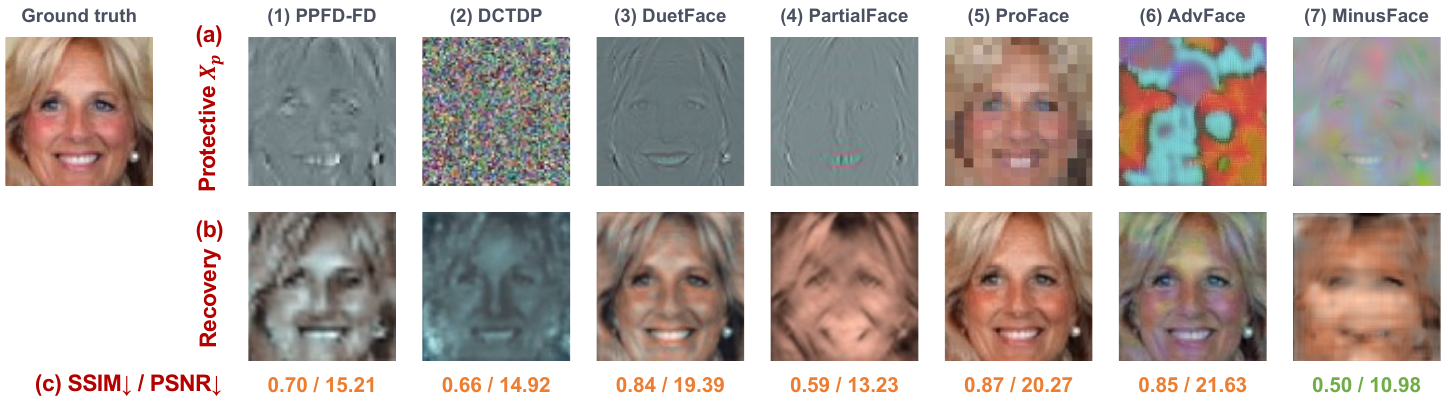}
   \caption{Privacy protection of MinusFace, compared with SOTAs. (a) MinusFace and most SOTAs successfully conceal the face image's visual appearance. (b) However, SOTAs fail to prevent recovery attacks. MinusFace outperforms all SOTAs as its recovered image is highly blurred and can hardly distinguish the face's existence. (c) Quantity results, where MinusFace exhibits the lowest SSIM and PSNR.} 
   \label{fig:protection}
   \vspace{-3mm}
\end{figure*}

\subsection{Recognition accuracy}
\label{subsec:exp-accuracy}

\noindent \textbf{Compared methods.} We compare MinusFace with an unprotected baseline and 9 transform-based PPFR methods. Specifically\footnote{We found no open-source code for PPFR-FD and AdvFace. We reproduce them to our best effort, recognizing the possibility of inconsistencies.}, (1) \textbf{ArcFace}~\cite{DBLP:conf/cvpr/DengGXZ19}, a non-privacy-preserving FR model trained on original face images; (2) \textbf{PEEP}~\cite{DBLP:journals/compsec/ChamikaraBKLC20}, which obfuscates images using differential noise (privacy budget set to 5); (3) \textbf{InstaHide}~\cite{DBLP:conf/icml/Huang0LA20}, mixing the face image with 2 other images to conceal appearance; (4) \textbf{Cloak}~\cite{DBLP:conf/www/MireshghallahTJ21}, compressing the image's feature space (trade-off parameter set to 100); (5) \textbf{PPFR-FD}~\cite{wang22ppfrfd}, shuffling and mixing frequency channels; (6) \textbf{DCTDP}~\cite{DBLP:conf/eccv/JiWHWXDZCJ22}, appending a frequency noise perturbation mask (privacy budget set to $\epsilon$=1); (7) \textbf{DuetFace}~\cite{DBLP:conf/mm/MiHJLXDZ22}, pruning frequency components and restoring accuracy via two-party collaboration; (8) \textbf{PartialFace}~\cite{mi2023privacy} exploiting a random subset of frequency channels for recognition; (9) \textbf{ProFace}~\cite{yuan2022pro}, hiding the image's appearance through deep steganography; (10) \textbf{AdvFace}~\cite{wang2023privacy}, perturbing the image by cyclically adding adversarial noise. These methods are divided into two branches by their means: the first three are early works that indiscriminately perturb all features, while the remaining selectively perturb the most visually informative features to better maintain accuracy.

\noindent \textbf{Performance analysis.} We evaluate MinusFace, baseline and compared methods on LFW, CFP-FP, AgeDB, CPLFW, and CALFW, and report results as recognition accuracy. We also evaluate them on IJB-B and IJB-C, and report TPR@FPR(1e-4). Results are summarized in~\cref{tab:comp-sota}.

We observe that early methods~\cite{DBLP:journals/compsec/ChamikaraBKLC20,DBLP:conf/www/MireshghallahTJ21,DBLP:conf/icml/Huang0LA20} experience a significant performance drop, especially on IJB-B/C, due to the compromise of identity features in indiscriminate obfuscation. Despite being designed to conceal mostly visual features, \cite{yuan2022pro,wang2023privacy} also exhibit considerable accuracy loss, suggesting inefficient trade-off between identity and visual features. Recently, frequency-based methods~\cite{wang22ppfrfd,DBLP:conf/eccv/JiWHWXDZCJ22,DBLP:conf/mm/MiHJLXDZ22,mi2023privacy} achieve notable accuracy by pruning visual appearance through removing low-frequency channels at a marginal utility cost. Their performance closely approaches the unprotected baseline. However, we later show that they can be susceptible to recovery attacks. MinusFace attains commendable performance, with a small gap ($\leq 2\%$) from the unprotected baseline. It is on par ($\pm 1\%$) with frequency-based methods and outperforms all other SOTAs. We argue that this slight accuracy trade-off is justified, as MinusFace offers significantly improved protection capability and efficiency, later discussed in~\cref{subsec:exp-recovery,subsec:exp-cost}.

\subsection{Concealing of visual information}
\label{subsec:exp-visual}

To evaluate MinusFace's privacy protection, recall our two-fold privacy goals: \textit{visually concealing the face's appearance} and \textit{hindering recovery attacks}. Here, we focus on the first goal and compare MinusFace with PPFR-FD, DCTDP, DuetFace, PartialFace, ProFace, and AdvFace. These SOTAs, similar to ours, treat visual and identity features discriminately. Specifically, we visualize their protective representations $X_p$ to determine if visual appearances can be discerned. Note that $X_p$ are not all in the form of images: Frequency-based methods produce $X_p$ as frequency channels, which we convert back via a reverse transform; AdvFace generates feature maps, transformed into images using its shadow model; ProFace directly creates images.

\Cref{fig:protection}(a) displays $X_p$ of each SOTA and MinusFace. Generally, all methods successfully conceal the face's appearance. DuetFace and ProFace provide slightly inferior protection, as their generated $X_p$ reveal some discernible facial features. DCTDP and AdvFace better conceal visual appearance by applying noise and obfuscation. In~\cref{fig:protection}a(7), MinusFace produces $X_p$ that nearly eliminates the face's structural clues and completely conceals its texture details, effectively achieving the first privacy goal.

\subsection{Protection against recovery}
\label{subsec:exp-recovery}

We here analyze the second privacy goal of protecting against recovery. We find MinusFace provides significantly better protection than SOTAs. We first describe the attack.

\noindent \textbf{Threat model.} We consider a white-box attacker who can query the PPFR framework and know its detailed protection mechanism. This attacker is typically envisioned as a malicious third-party wiretapping the transmission. While aware of the framework's general setup, such as hyper-parameters, the attacker does not know the specific sample-wise parameters (\eg, $\theta$ in our case) used by the client to generate protective representations $X_p$. Assume the attacker has access to a training dataset of face images $X$. It can first obtain $X_p$ by querying the PPFR framework. Then, it can train a recovery model $f^{-1}$ to map $X_p$ back to $X$, as $\argmin_{\delta}~\left\|f^{-1}(X_p;\delta),X\right\|_{1}$, and exploit $f^{-1}$ to recover the client's shared $X_p$. We concretely use BUPT~\cite{DBLP:journals/corr/abs-1911-10692} of 1.3M images as the attacker's dataset, and employ a full-scale U-Net~\cite{DBLP:conf/miccai/RonnebergerFB15} as its $f^{-1}$. 

\noindent \textbf{Comparison with SOTAs.} We train an attack model for MinusFace and each SOTA. \Cref{fig:protection}(b) displays examples of recovered images. We find that most prior arts provide insufficient protection against recovery. Specifically, ProFace is not designed to prevent recovery, resulting in a faithful recovered image from its protective $X_p$ in~\cref{fig:protection}b(5). Some prior arts~\cite{wang22ppfrfd,DBLP:conf/mm/MiHJLXDZ22,wang2023privacy} suggest resistance to the attack. However, in~\cref{fig:protection}b(1)(3)(6), we find they can actually be recovered by a more powerful attacker, \eg, training $f^{-1}$ on a much larger dataset. For other methods, the faces' appearance can also be somewhat recovered. We attribute the SOTAs' shortcomings to their setback in ensuring adequate removal of visual features, especially facial textures. This leaves potential features that attackers can leverage. In~\cref{fig:protection}b(7), MinusFace overcomes the drawback and exhibits strong protection to recovery, outperforming SOTAs.

\noindent \textbf{Quantitative comparison.} In~\cref{fig:protection}(c), we quantify the quality of recovered images by their structural similarity (SSIM) and peak signal-to-noise ratio (PSNR) compared to the ground truth. Results are averaged on 10K IJB-C images. MinusFace exhibits the lowest SSIM and PSNR, indicating the best protection.

\subsection{Protection against dedicated attacks}
\label{subsec:exp-recovery-2}

\begin{figure}[tbp]
  \centering
   \includegraphics[width=\linewidth]{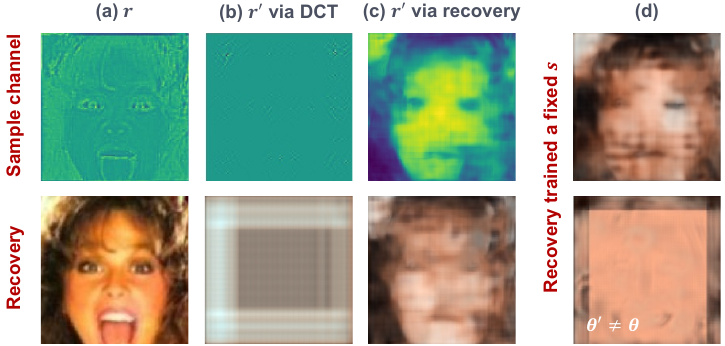}
   \caption{Left: Sample channels from $r$ and the attacker's attempted inversions to reproduce $r'$, together with their recovery. (a) $r$ is not designed as privacy-preserving, hence can be recovered. (b-c) However, the attacker cannot obtain $r$ or its correct inversion $r'$, making recovery infeasible. Right: (d) Training recovery model on fixed $\theta$ does not pose an effective threat, as it fails entirely for $\theta'\neq\theta$, where $\theta$ has a random space of $192!$.}
   \label{fig:recovery-2}
   \vspace{-4mm}
\end{figure}

We further investigate two attacks dedicated to MinusFace's design that attempt to invert $r$ or bypass $X_p$'s randomness. 

\noindent \textbf{Inverting $r$.} It is crucial to note that the high-dimensional residue $r$ is not designed to be protective (although it produces a protective $X_p$) and can be easily recovered (\cref{fig:recovery-2}(a)). Our design is safe as $r$ is never shared. Yet, an attacker may attempt to further invert $r$ from $X_p$ and carry out recovery therefrom. However, we demonstrate that \textit{inverting $r$ is infeasible}. First, the attacker cannot invert an $r'=r$ by re-encoding it as $r'=e(X_p)$, even if it knows the specific $\theta$ (we omit $\theta$ for simplicity). Note that although~\cref{eq:mapping-property} assures $d(e(X))=X$, its opposite $e(d(x))=x$ is not guaranteed to hold. In~\cref{fig:recovery-2}(b), re-encoding $X_p$ produces inconsistent and uninformative $r'$, further demolishing the attack. The attacker also cannot train a recovery model from $X_p$ to $r$ as it is essentially as difficult as training the previously discussed $f^{-1}$. \Cref{fig:recovery-2}(c) demonstrates the unsuccessful $r'$ and its recovery.

\noindent \textbf{Bypassing randomness}. The attacker is capable of generating $X_p$ under a specific shuffle seed $\theta$. It hence can train $f^{-1}$ on $X_p$ from the same $\theta$, to bypass the randomness of representations. In~\cref{fig:recovery-2}(d), such trained $f^{-1}$ produces slightly better recovery on $X_p$ under the same $\theta$, but fails entirely for any $\theta'\neq \theta$. As $\theta$ has a total random space of $192!$, this attack does not impose any effective threat.

\subsection{Ablation study}
\label{subsec:exp-abal}

\begin{table}[tbp]
\centering
\begin{tabular}{lccc}
\toprule
\textbf{Method}         & \textbf{CFP-FP}                    & \textbf{AgeDB}                     & \textbf{CPLFW}                     \\
\midrule
ArcFace        & 98.30                     & 97.88                     & 92.77                     \\
$r$          & 98.27                     & 97.82                     & 92.73                     \\
$R'$ & 53.85   & 54.71 & 51.20 \\
$X_p$ \textbf{(default)} & 96.92                          & 97.57                          & 91.90                          \\
\bottomrule
\end{tabular}
\caption{Accuracy of models trained alternatively on $r$ and $R'$.}
\label{tab:abal}
\vspace{-2mm}
\end{table}

\noindent \textbf{Recognition accuracy of $r$ and $R'$.} Recall $r$ and $R'$ represent the high-dimensional residue and its spatial decoding, respectively. We train FR models on them and show their recognition accuracy. We expect the model trained on $r$ (\ie, $f$) to achieve high accuracy, as it indicates a lossless feature subtraction, favorable for MinusFace's overall utility. We also expect $R'$'s model to experience a significant accuracy downgrade (close to a random guess of $50\%$), as $R'$ should be fully removed of features. Results in~\cref{tab:abal} meet our expectations. We yield further ablation studies to the supplementary material, due to the limit of space.

\subsection{Efficiency and compatibility}
\label{subsec:exp-cost}

\begin{table}[tbp]
\begin{tabular}{l|cccccc}
\toprule
\textbf{Method}  & ours & \cite{wang22ppfrfd} & \cite{DBLP:conf/eccv/JiWHWXDZCJ22} & \cite{DBLP:conf/mm/MiHJLXDZ22} & \cite{mi2023privacy}  & \cite{wang2023privacy} \\
\midrule
\textbf{Storage} $\downarrow$ & 1   & $\times$36  & $\times$63  & $\times$54  & $\times$9 & $\times$5.3 \\
\bottomrule
\end{tabular}
\caption{Comparison of storage and transmission cost. $\times n$ indicates an $n$-time larger protective representation than MinusFace. }
\label{tab:storage}
\vspace{-3mm}
\end{table}

A practical PPFR framework is expected to have low inference latency and be efficient for storage and transmission. MinusFace is an ideal fit for these goals.

\noindent \textbf{Latency.} Testing on a personal laptop, MinusFace costs an average of $69$ ms to turn an image into protective $X_p$, an order of magnitude smaller than typical communication time. It does not increase time costs for the service provider.

\noindent \textbf{Storage and transmission.} Many prior arts~\cite{wang22ppfrfd,DBLP:conf/eccv/JiWHWXDZCJ22,DBLP:conf/mm/MiHJLXDZ22,mi2023privacy,wang2023privacy} produce $X_p$ as high-dimensional feature channels (\eg, DCTDP generates an $(189,H,W)$ $X_p$), resulting in additional storage and transmission costs. In contrast, MinusFace produces spatial images. As shown in~\cref{tab:storage}, MinusFace's $X_p$ requires far less size compared to SOTAs.

\noindent \textbf{Compatibility.} MinusFace is also compatible with different SOTA FR backbones and training objectives, as it neither modifies nor requires any specific design of FR architecture.

\section{Conclusion}
\label{sec:conclusion}

This paper investigates the privacy protection of face images. We present a new methodology, \textit{feature subtraction}, to generate privacy-preserving face representations by capturing the residue between an original face and its regeneration. We further ensure the recognizability and privacy of residue via high-dimensional mapping and random channel shuffling, respectively. Our findings are concretized into a novel PPFR method, MinusFace. Extensive experiments demonstrate that it achieves satisfactory recognition accuracy and enhanced privacy protection.

{
    \small
    \bibliographystyle{ieeenat_fullname}
    \bibliography{main}
}

\clearpage
\setcounter{page}{1}
\setcounter{section}{0}
\renewcommand{\thesection}{\Alph{section}} 
\maketitlesupplementary

\noindent This supplementary material provides additional details on the following about the proposed MinusFace method:

\begin{itemize}[leftmargin=2em]
    \item Experimental setup and implementation details;
    \item Further methodological and experimental discussions;
    \item Further ablation studies;
    \item Additional image visualization;
    \item Ethics discussion.
\end{itemize}

\section{Detailed experimental setup}
\label{sec:supp-exp-setup}

This section provides information about our experimental setup. We first further introduce the employed datasets and backbones, then discuss our detailed implementations.

\subsection{Datasets}

\noindent \textbf{Training datasets.} We train our FR models on the widely-used MS1Mv2 dataset~\cite{DBLP:conf/eccv/GuoZHHG16}, which consists of 5.8M face images from 85K distinct individuals, mainly celebrities. In accordance with CVPR guidelines, we provide further ethical discussion in~\cref{sec:supp-ethics}. Additionally, we utilize the smaller BUPT-BalancedFace dataset~\cite{DBLP:journals/corr/abs-1911-10692} (BUPT) for recovery attacks, comprising 1.3M images from 28K identities.

\noindent \textbf{Test datasets.} We compare MinusFace and SOTA methods on 7 datasets. (1) We benchmark on five widely-used, regular-sized datasets and report results as test accuracy (with a lower bound of 50\%): LFW~\cite{LFWTech}, 13K web-collected images from 5.7K individuals; CALFW~\cite{DBLP:journals/corr/abs-1708-08197} and CPLFW~\cite{CPLFWTech}, reorganized version of LFW that enhances cross-age and cross-pose variations, respectively; AgeDB~\cite{DBLP:conf/cvpr/MoschoglouPSDKZ17} and CFP-FP~\cite{DBLP:conf/wacv/SenguptaCCPCJ16}, similar in size and varied in age and pose. Notably, LFW is often viewed as a saturated dataset, where the highest recognition accuracy is anticipated. Conversely, CFP-FP and CPLFW pose greater challenges to SOTA FR methods due to their increased pose variations. (2) We extend our study to two large-scale benchmarks: IJB-B~\cite{DBLP:conf/cvpr/WhitelamTBMAMKJ17} and IJB-C~\cite{DBLP:conf/icb/MazeADKMO0NACG18}, which provide 80K and 150K still images and video frames, respectively. Results are reported as TPR@FPR(1e-4), \ie, the true positive rate (TPR) at a specific false positive rate (FPR) of 1e-4.

\subsection{Backbones}

We employ an adapted ResNet50 model with an improved residual unit (IR-50)~\cite{DBLP:conf/cvpr/HeZRS16} as the FR backbone for both $f$ and $f_p$. The only modification to the backbone is changing the input channels of $f$ to 192 to match the number of channels in $x$. For the generative model $g$ and the attacker's recovery model $f^{-1}$, we utilize U-Net~\cite{DBLP:conf/miccai/RonnebergerFB15}, a popular architecture for image generation and segmentation, which can be considered an autoencoder with skip connections. We employ a full-scale U-Net for $f^{-1}$ to enhance the attacker's capability and a smaller one for $g$. The latter helps us to reduce MinusFace's local storage and inference time.

\subsection{Implementation details}

\noindent \textbf{Image preprocessing.} We preprocess the training and test datasets using standard methods: We crop faces from the images and align their positions based on the 5-point landmarks of the faces (positions of eyes, nose, and lips). To improve the FR model's generalization, we apply random horizontal flips to the training images.

\noindent \textbf{Encoding and decoding mappings.} We opt for DCT and its inverse IDCT as our concrete encoding and decoding mappings. DCT is a popular spatial-frequency transformation first employed by the JPEG compression standard~\cite{DBLP:journals/cacm/Wallace91}. It converts a $(3,H,W)$ spatial image into $(192,H,W)$ frequency channels. Specifically, the spatial image first undergoes an 8-fold up-sampling to obtain a shape of $(3,8H,8W)$. As DCT later divides $H$ and $W$ by 8, this makes sure the resulting frequency channels have consistent shapes. Subsequently, each channel of the image is split into $(8,8)$-pixel blocks. DCT turns each block into a 1D array of 64 frequency coefficients and reorganizes all coefficients from the same frequency across blocks into an $(H, W)$ frequency channel, that is spatially correlated to the original $X$. This conversion produces 64 frequency channels from each of the 3 spatial channels. These channels are then stacked to form the final shape of $(192, H, W)$. We further discuss the properties of DCT in~\cref{subsec:supp-discuss-dct}.

\noindent \textbf{Training.} Our training involves two stages: First to produce $r$, we train the generative model $g$ and the FR model $f$ in an end-to-end manner. Then, we freeze $g$ and train $f_p$ on $X_p$, which is generated from the decoding of randomly shuffled $r$. For both stages, we train the model from scratch for 24 epochs with a stochastic gradient descent (SGD) optimizer. We choose 64, 0.9, and 1e-4 for batch size, momentum, and weight decay, respectively. The training of $f,f_p$ starts with an initial learning rate of 1e-2, which is successively divided by a factor of 10 at epochs 10, 18, and 22. The learning rate of $g$ is further halved (\ie, starting at 5e-3) since the generative model requires a smaller learning rate to facilitate convergence. We choose the weights for our training objective as $\alpha=5,\beta=1$. To help $f_p$ generalize on randomized representation, we augment the dataset 3 times, similar to~\cite{mi2023privacy}, meaning each $X$ generates three $X_p$ from distinct random shuffling. We apply the same training settings for SOTAs. For the recovery attacker, we train its model until convergence using an initial learning rate of 1e-3. Experiments are carried out in parallel on 8 NVIDIA Tesla V100 GPUs with PyTorch 1.10 and CUDA 11. We use the same random seed for all experiments.

\section{Further discussion}
\label{sec:supp-discuss}

\subsection{Properties of DCT}
\label{subsec:supp-discuss-dct}

\noindent \textbf{DCT is invertible.} DCT and IDCT together are invertible as they, by design, map a spatial image to and from its frequency channels \textit{losslessly}. Hence, it naturally holds:

\begin{equation}
    \label{eq:supp-inv-rewrite}
    d(e(X)) = X.
\end{equation}

\noindent Importantly, it however does not guarantee

\begin{equation}
    e(d(x)) = x,
\end{equation}

\noindent unless $x$ is a \textit{valid frequency representation} that is directly encoded via DCT from a spatial image $X$ as $x=e(X)$ (this case satisfies~\cref{eq:x-invertible}). In MinusFace, since $x',r$ are not encoded via DCT but derived from regeneration and feature subtraction, they are not frequency representations of $X', R'$ but rather serve as generic high-dimensional representations. Therefore, we can expect

\begin{equation}
    e(d(x')) \neq x',
    e(d(r)) \neq r.
\end{equation}

\noindent This property benefits privacy, as an attacker cannot invert a plausible $r$ from $X',R'$ or $X_p$. We previously demonstrated this in~\cref{subsec:exp-recovery-2}.

\noindent \textbf{DCT is homomorphic.} DCT is a linear transformation. Any linear transformation is additively homomorphic by definition. Hence it satisfies

\begin{equation}
    d(x_1+x_2)= d(x_1)+d(x_2).
\end{equation}

\noindent \textbf{DCT is used differently by MinusFace and SOTAs.} Notably, some SOTA PPFR methods~\cite{wang22ppfrfd,DBLP:conf/eccv/JiWHWXDZCJ22,DBLP:conf/mm/MiHJLXDZ22,mi2023privacy} also utilize DCT in their pipelines. Both MinusFace and SOTAs are likely inspired by~\cite{DBLP:conf/cvpr/0007QSWCR20}, which demonstrates that an image recognition model can perform accurately on DCT components, making DCT an ideal choice for lossless transformation. However, their motivations differ: SOTAs utilize DCT to exploit specific properties of frequency representations (the perceptual disparity among frequency channels), which allows for heuristic channel pruning. On the other hand, MinusFace employs DCT for its invertible and homomorphic mapping properties, as well as its high-dimensional redundancy, which helps to produce identity-informative $r$ and privacy-preserving $R'$.

\noindent \textbf{DCT is replaceable with other mappings.} As discussed in~\cref{subsec:high-dim-constrain}, DCT/IDCT can be replaced by other mapping algorithms that satisfy~\cref{eq:mapping-property}. We present an alternative option, discrete wavelet transform (DWT), in~\cref{subsec:supp-abal-mapping}. Nonetheless, we empirically find that DCT offers a better trade-off between accuracy and privacy.

\subsection{Recovery result of SOTAs}
\label{subsec:supp-discuss-recovery}

Some prior studies~\cite{wang22ppfrfd,DBLP:conf/mm/MiHJLXDZ22,DBLP:conf/eccv/JiWHWXDZCJ22,wang2023privacy} claim that their proposed methods are resistant to recovery attacks. However, in~\cref{fig:protection}(b), this paper finds that they can, to some extent, be recovered. We investigate the cause of these distinct experimental outcomes. For the attacker, successful recovery depends on various factors, such as training resources (\eg, the volume of training data) and strategy (\eg, choice of training objective, optimizer, learning rate, and batch size). In fact, this paper examines \textit{a more advanced attacker} than those in SOTAs. 

\noindent \textbf{Our attacker exploits a larger training dataset.} Among SOTAs, PPFR-FD and DuetFace employ a recovery model trained on just $\leq$100K face images. AdvFace trains its recovery model on 500K images. In contrast, our analyzed attacker can exploit the entire BUPT dataset, which consists of 1.3M images. The increased data volume can plausibly enhance the attacker's capability.

\noindent \textbf{Our attacker employs an improved training strategy.} For recovery, we empirically find that selecting appropriate learning rates and batch sizes is crucial. DCTDP opts for a learning rate of 1e-1 and a batch size of 512, which could be too large for the recovery model to stably converge. In this paper, we choose a learning rate of 1e-3 and a batch size of 64, which we find facilitate recovery.

Despite the advancement in the attacker's capability, MinusFace still effectively prevents the successful recovery of protected faces. This further demonstrates the robust protection provided by MinusFace.

\section{Further ablation study}
\label{sec:supp-abal}

\subsection{Choice of FR backbones and objectives}
\label{subsec:supp-abal-fr}

\begin{table}[tbp]
\centering
\begin{tabular}{lccc}
\toprule
\textbf{Method}               & \textbf{CFP-FP} & \textbf{AgeDB} & \textbf{CPLFW} \\
\midrule
IR-18, unprotected   & 92.31                      & 94.65 & 89.41 \\
IR-18, MinusFace     & 90.21                      & 93.25 & 87.60 \\
\midrule
CosFace, unprotected & 92.89                      & 95.15 & 89.52 \\
CosFace, MinusFace   & 89.23  & 94.77 & 86.97 \\
\bottomrule
\end{tabular}
\caption{Compatibility of MinusFace. Combining MinusFace with different FR backbones (IR-18) or losses (CosFace) also sustains high accuracy, compared to their unprotected baselines.}
\label{tab:supp-compatibility}
\end{table}

As discussed in~\cref{subsec:exp-cost}, MinusFace is compatible with different SOTA FR backbones and training objectives. To illustrate this, we combine MinusFace with a distinct IR-18 FR model and CosFace~\cite{DBLP:conf/cvpr/WangWZJGZL018} loss, using BUPT as the training dataset. \Cref{tab:supp-compatibility} shows the recognition accuracy on CFP-FP, AgeDB, and CPLFW, comparing MinusFace with unprotected baselines. MinusFace maintains a stable performance that is close to the unprotected baseline.

\subsection{Choice of encoding and decoding mappings}
\label{subsec:supp-abal-mapping}

\begin{figure}[tbp]
  \centering
   \includegraphics[width=\linewidth]{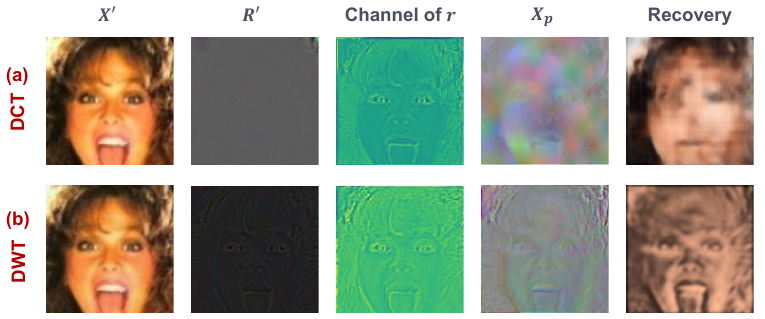}
   \caption{Comparison between DCT and DWT. We replace our mappings with DWT and its inverse and visualize $X',R'$, sample channel of $r$, $X_p$, and its recovery for DCT and DWT, respectively. DWT also succeeds in creating an almost blank $R'$ and visually indiscernible $X_p$, illustrating the effectiveness of MinusFace. However, the attacker can achieve a clearer recovery on DWT, making DCT a more secure choice.}
   \label{fig:supp-dwt}
\end{figure}

\begin{table}[tbp]
\centering
\begin{tabular}{lccc}
\toprule
\textbf{Method}               & \textbf{CFP-FP} & \textbf{AgeDB} & \textbf{CPLFW} \\
\midrule
MinusFace (default)     & 90.21                      & 93.25 & 87.60 \\
DWT   & 90.47                      & 93.53 & 87.92 \\
DCT, masking   & 81.40                      & 88.03 & 82.82 \\
\bottomrule
\end{tabular}
\caption{Ablation studies of MinusFace. Replacing DCT with DWT achieves comparable recognition accuracy. Yet, replacing shuffling with random masking significantly degrades accuracy.}
\label{tab:supp-abal}
\end{table}

To demonstrate a generic pair of encoding and decoding satisfying~\cref{eq:mapping-property} may also serve as our mappings $e,d$, We replace DCT/IDCT with an alternative choice: discrete wavelet transform (DWT). By default, DWT converts the $(3,H,W)$ image into a $(12,H,W)$ representation. We compare its recognition accuracy, concealment of visual images, and protection against recovery to default DCT.

\Cref{fig:supp-dwt} demonstrates the outcomes using DWT and its inverse (IWT) as the mappings, compared with those using DCT/IDCT, respectively. It can be observed that MinusFace effectively produces almost blank $R'$ and visually uninformative $X_p$ for both DCT and DWT. \Cref{tab:abal} further demonstrates that replacing DCT with DWT can achieve on-par recognition accuracy. However, we find that $X_p$ generated via DWT is less resistant to recovery, as face contours can still be observed in the recovered image. We attribute this relative deficiency to two reasons: (1) $X_p$ generated from DWT is less obfuscated in texture details, as its shuffling is less randomized, and (2) DWT produces a significantly smaller random space of $12!$ ($192!$ as of DCT), which eases the attacker's learning of consistent representations.

The results indicate that DCT can indeed be replaced with other mappings. However, the specific choice of $d,e$ should also be more carefully considered based on the experimental context.

\subsection{Choice of perturbation on $r$}
\label{subsec:supp-abal-perturb}

\begin{figure}[tbp]
  \centering
   \includegraphics[width=\linewidth]{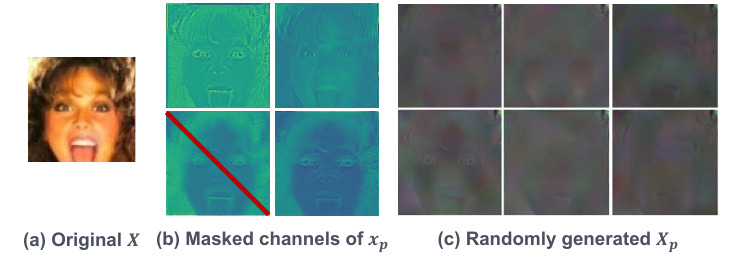}
   \caption{Replacing random shuffling with (b) masking also results in (c) $X_p$ revealing slightly recognizable features at a marginal cost to privacy, which aligns with our expectations.}
   \label{fig:supp-masking}
\end{figure}

In~\cref{subsec:rand-shuffle}, we opt for random channel shuffling as our perturbation. We here demonstrate it achieves better privacy and accuracy trade-off than other options. Specifically, we generate $r$ as usual yet replace $s(r;\theta)$ with random \textit{masking} of channels $m(r;\theta)$. We produce $X_p=d(m(r;\theta))$, where $\theta$ denotes the random seed. We choose a masking ratio of $25\%$. \Cref{fig:supp-masking} (b) shows its process.

As per the derivation in~\cref{eq:perturb}, $X_p=d(m(r;\theta))$ should also reveal recognizable features of $X$ at marginal costs to privacy. We observe that the produced $X_p$ in~\cref{fig:supp-masking} aligns with our expectations. However, we find that it suffers from a downgrade in recognition accuracy, as shown in~\cref{tab:supp-abal}, since the features it reveals could be too subtle for FR models to effectively leverage. In this sense, random channel shuffling better balances privacy and accuracy.

\subsection{Without feature subtraction}
\label{subsec:supp-abal-wofs}

\begin{figure}[tbp]
  \centering
   \includegraphics[width=0.95\linewidth]{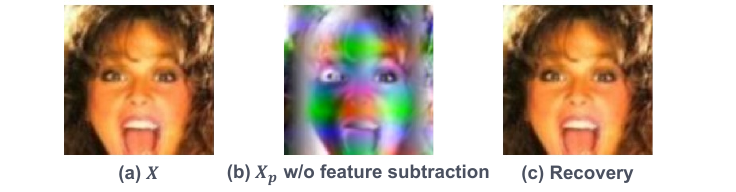}
   \caption{Ablation study of feature subtraction. (b) $X_p$ is directly produced from the high-dimensional representation of (a) $X$. It reveals clear visual features and is easy to (c) be recovered.}
   \label{fig:supp-wofs}
\end{figure}

We discuss the importance of feature subtraction in achieving privacy protection. The primary outcome of feature subtraction is to create a recognizable $r$ that precisely maps to a blank $R'$. By perturbing $r$, we factually \textit{restore identity features} in $R'$ to produce $X_p$ with minimum privacy cost. Here, we emphasize the critical role of first achieving a blank $R'$.

To ablate feature subtraction, we directly perform random channel shuffling on $x$ (\ie, the high-dimensional representation of $X$) to obtain $X_p$. \Cref{fig:supp-wofs} illustrates the resulting $X_p$ and its recovery, where privacy is completely undermined as $X_p$ contain a wealth of visual features. Recall that~\cref{eq:perturb} is based on the premise that $d(r)\rightarrow 0$. Without feature subtraction, this condition would not be satisfied, as $R'=d(r)$ would not be blank.

\section{Additional example images}

\begin{figure}[tbp]
  \centering
   \includegraphics[width=\linewidth]{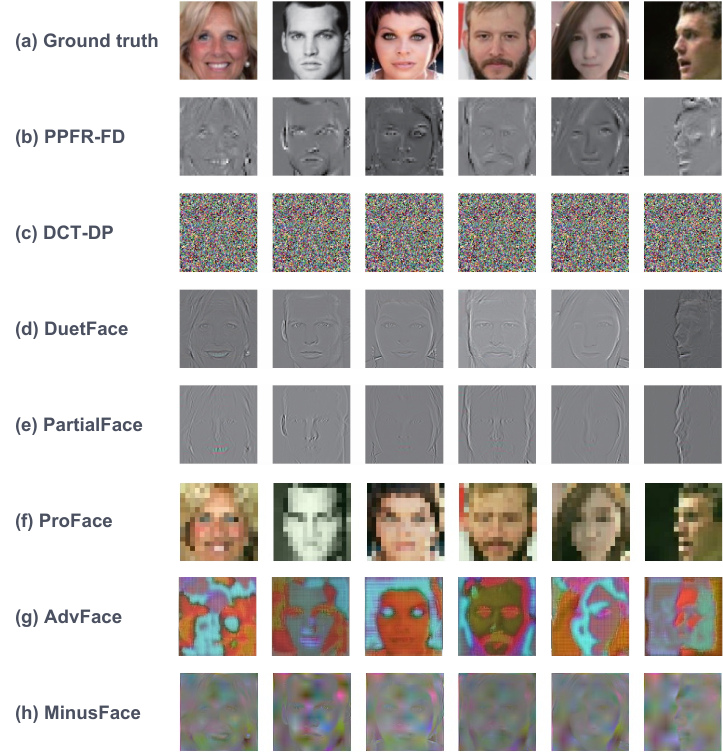}
   \caption{Additional sample images for the protective $X_p$.}
   \label{fig:supp-xs}
   \vspace{-2mm}
\end{figure}

\begin{figure}[tbp]
  \centering
   \includegraphics[width=\linewidth]{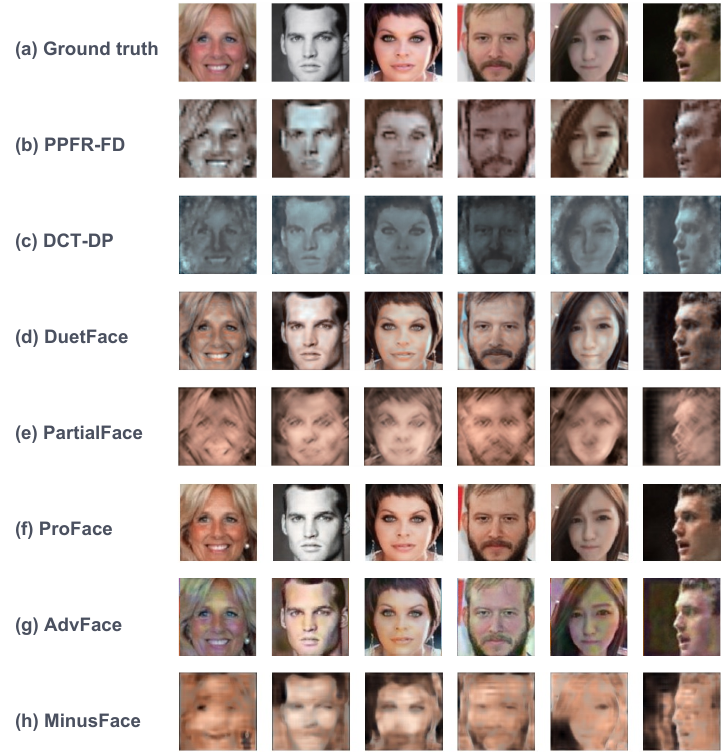}
   \caption{Additional sample images for the recovery from $X_p$.}
   \label{fig:supp-recovery}
  \vspace{-2mm}
\end{figure}

To demonstrate the generality of MinusFace, we further supplement~\cref{fig:protection} with additional example images for the protective representation $X_p$ and its recovery, comparing MinusFace with SOTAs. Specifically,~\cref{fig:supp-xs} illustrates $X_p$ and~\cref{fig:supp-recovery} illustrates the recovery.

\section{Ethics discussion}
\label{sec:supp-ethics}

Our models primarily utilize the MS1Mv2 dataset, a modified version of the MS-Celeb-1M (MS1M) dataset provided by the InsightFace project\footnote{https://github.com/deepinsight/insightface/}, containing celebrity face images. As personal characteristics such as facial semantics may be inferred from the dataset, we are obliged to justify its use per CVPR ethics guidelines. The reasons for using MS1Mv2 include its essential role in ensuring fair comparisons: It is one of the \textit{de facto} standard training datasets in face recognition, and is employed by the majority of the methods~\cite{DBLP:conf/cvpr/DengGXZ19,DBLP:journals/compsec/ChamikaraBKLC20,DBLP:conf/www/MireshghallahTJ21,DBLP:journals/tifs/TsengW20,DBLP:conf/mm/MiHJLXDZ22,DBLP:conf/eccv/JiWHWXDZCJ22,mi2023privacy} we compare.

\end{document}